\documentclass[11pt,a4paper]{article}
\usepackage[hyperref]{acl2018}
\usepackage{times}
\usepackage{latexsym}

\usepackage{url}
\usepackage[utf8]{inputenc}
\usepackage{comment}
\usepackage{graphicx}
\usepackage{url}
\usepackage{multirow}

\aclfinalcopy 
\def\aclpaperid{453} 

\newcommand{\MC}{\multicolumn}
\newcommand{\MR}{\multirow}

\title{Filtering and Mining Parallel Data in a Joint Multilingual Space}

\author{Holger Schwenk \\
  Facebook AI Research \\
  {\tt schwenk@dfb.com} \\ \\} 

\date{}

\begin{document}

\maketitle \begin{abstract}
We learn a joint multilingual sentence embedding and use the distance between
sentences in different languages to filter noisy parallel data and to mine for
parallel data in large news collections.  We are able to improve a competitive
baseline on the WMT'14 English to German task by 0.3 BLEU by filtering out 25\%
of the training data.  The same approach is used to mine additional bitexts for
the WMT'14 system and to obtain competitive results on the BUCC shared task to
identify parallel sentences in comparable corpora.

The approach is generic, it can be applied to many language pairs and it is
independent of the architecture of the machine translation system.
\end{abstract}

\section{Introduction}

Parallel data, also called bitexts, is an important resource to train neural
machine translation systems (NMT).  It is usually assumed that the quality of
the automatic translations increases with the amount of available training
data.  However, it was observed that NMT systems are more sensitive to
noise than SMT systems, e.g. \citep{belinkov:2017:noise_in_nmt}.
Well known sources of parallel data are international organizations like the
European Parliament or the United Nations, or community provided translations
like the TED talks.  In addition, there are many texts on the Internet which
are potential mutual translations, but which need to be identified and
aligned.  Typical examples are Wikipedia or news collections which report on
the same facts in different languages.  These collections are usually called
comparable corpora.

In this paper we propose an unified approach to filter noisy bitexts and to
mine bitexts in huge monolingual texts.  The main idea is to first learn
a joint multilingual sentence embedding. 
Then, a threshold on the distance between two sentences in this joint embedding space
can be used to filter bitexts (distance between source and target sentences),
or to mine for additional bitexts (pairwise distances between all source and
target sentences). No additional features or classifiers are needed.

\section{Related work}
\label{SectRelated}

The problem of how to select parts of bitexts has been addressed before, but
mainly from the aspect of domain adaptation
\citep{axelrod-he-gao:2011:EMNLP,santamaria:2017:iswlt_nmtdsel}.  It was
successfully used in many phrase-based MT systems, but it was reported to be
less successful for NMT \citep{wees:2017:emnlp_nmtdsel}.
It should be stressed that domain adaptation is different from filtering noisy
training data.  Data selection extracts the most relevant bitexts for the test
set domain, but does not necessarily remove wrong translations, e.g. source and
target sentences are both in-domain and well formed, but they are not mutual
translations.

There is a huge body of research on mining bitexts, e.g. by analyzing
the name of WEB pages or links \citep{resnik:2003:cl_web}.  Another
direction of research is to use cross-lingual information retrieval, e.g.
\citep{utiyama:2003:acl,Munteanu:2005:cl_bitexts,Rauf:2009:eacl}.
There are some works which use joint embeddings in the process of filtering or
mining bitexts.
For instance, \citet{bucc:2017:eval_rali} first embed sentences into two
separate spaces.  Then, a classifier is learned on labeled data to decide
whether sentences are parallel or not.  Our approach clearly outperforms this
technique on the BUCC corpus (cf. section~\ref{SectBucc}).
\citet{bucc:2018:eval_cmu} use averaged multilingual word embeddings to
calculate a joint embedding of all sentences. However, distances between all
sentences are only used to extract a set of potential mutual translation. The
decision is based on a different system.
In \citet{msr:2018:arxiv_enzh} NMT systems for Zh~$\leftrightarrow$~En are
learned using a joint encoder. A sentence representation is obtained as the
mean of the last encoder states. Noisy bitexts are filtered based on the distance.
In all these works, embeddings are learned for two languages only, while we learn
one joint embedding for up to nine languages.

\section{Multilingual sentence embeddings}

We are aiming at an embedding of entire sentences in different languages into
one joint space, with the goal that the distance in that space reflects their
semantic difference, independently of the language.
There are several works on learning multilingual sentence embeddings which
could be used for that purpose, i.e.
\citep{Hermann:2014:acl_multling,Pham:2015:multling,Zhou:2016:acl_crossling,Chandar:2013:multlin_deep,Mougadala:2016:naacl_biword}.

In this paper, we extend our initial approach \citep{Schwenk:2017:repl4nlp}.
The underlying idea is to use multiple sequence encoders and decoders and to
train them with $N$-way aligned corpora from the MT community.
Instead of using one encoder for each language as in the original paper, we use
a shared encoder which handles all the input languages.  Joint encoders (and
decoders) have already been used in NMT \citep{google:2016:arxiv_zeronmt}.  In
contrast to that work, we do not use a special input token to indicate the
target language.  Our joint encoder has no information at all on the encoded
language, or what will be done with the sentence representation.

We trained this architecture on nine languages\footnote{en, fr, es, it, pt, de,
da, nl and fi} of the Europarl corpus with about 2M sentences each. We use BPE
\citep{sennrich:2016:acl_bpe} to learn one 20k joint vocabulary for all the nine
languages.\footnote{Larger vocabularies achieve only slight improvements.}
The joint encoder is a 3-layer BLSTM.  The word embeddings are of size 384 and
the hidden layer of the BLSTM is 512-dimensional. The 1024 dimensional sentence
embedding is obtained by max-pooling over the BLSTM outputs.  Dropout is set to
0.1.  These settings are identical to those reported in
\citep{Schwenk:2017:repl4nlp}, with the difference that we observe slight
improvement by using a deeper network for the joint encoder.  Once the system
is learned, all the BLSTM decoders are discarded and we only use the
multilingual BLSTM encoder to embed the sentences into the joint space.

A very similar approach was also proposed in
\citet{epsana:2017:ieee_embed_mine}.  A joint NMT system with attention is
trained on several languages pairs, similar to
\citep{google:2016:arxiv_zeronmt}, including a special token to indicate the
target language.  After training, the sum of the encoder output states is used
to obtain a fixed size sentence representation.

\section{Experimental evaluation: BUCC shared task on mining bitexts}
\label{SectBucc}

\begin{table}[b!]
  \centering
  \begin{tabular}[t]{|l|*{3}{@{\,\,}r@{\,\,}}|*{2}{@{\,\,}r@{\,\,}}|}
  \hline
  Lang.& \MC{3}{@{\,\,}c|@{\,\,}}{Train} & \MC{2}{@{\,\,}c|}{Test} \\
  Pair & en & other & aligned & en & other \\
  \hline
  en-de & 400k & 414k &  9580 & 397k & 414k \\
  en-fr & 370k & 272k &  9086 & 373k & 277k \\
  en-ru & 558k & 461k & 14435 & 566k & 457k \\
  en-zh &  89k &  95k &  1899 &  90k &  92k \\
  \hline
  \end{tabular}
  \caption{BUCC evaluation to mine bitexts. Number of sentences and
	size of the gold alignments.}
  \label{TabBuccData}
\end{table}

Since 2017, the workshop on Building and Using Comparable Corpora (BUCC) is
organizing a shared task to evaluate the performance of approaches to mine for
parallel sentences in comparable corpora \citep{bucc:2018:eval}.
Table~\ref{TabBuccData} summarizes the available data, and Table~\ref{TabBuccRef}
the official results.
Roughly a 40th of the sentences are aligned.
The best performing system \textit{``VIC''} is based on the so-called STACC
method which was shown to achieve state-of-the-art performance
\citep{azpeita:2016:acl_stacc}. It combines probabilistic dictionaries, search
for similar sentences in both directions and a decision module which explores
various features (common word prefixes, numbers, capitalized true-case tokens,
etc). This STACC system was improved and adapted to the BUCC tasks with a word
weighting scheme which is optimized on the monolingual corpora, and a named
entity penalty.  This task adaption substantially improved the generic STACC
approach \citep{bucc:2018:eval_vic}.
The systems RALI \citep{bucc:2017:eval_rali} and H2 \citep{bucc:2018:eval_cmu}
have been already described in section~\ref{SectRelated}.  NLP2CT uses a
denoising auto-encoder and a maximum-entropy classifier
\citep{bucc:2018:eval_nlp2ct}.

\begin{table}[t!]
  \centering
  \begin{tabular}[t]{|l|*{4}{c|}}
  \hline
  System & en-fr & en-de & en-ru & en-zh \\
  \hline
    VIC'17  & 79 &  84 &   - &  - \\
    RALI'17 & 20 &   - &   - &  - \\
   LIMSI'17 &  - &   - &   - &  43 \\
  \hline
   VIC'18   & 81 & 86 & 81 & 77 \\
    H2'18   & 76 &  - &  - &  - \\
  NLP2CT'18 &  - &  - &  - & 56 \\
  \hline
  \hline
  \end{tabular}
  \caption{Official test set results of the 2017 and 2018 BUCC shared tasks (F-scores).}
  \label{TabBuccRef}
  \vspace{-10pt}
\end{table}

\begin{table}[b!]
  \centering
  \begin{tabular}[t]{|cl|*{4}{r}|}
  \hline
  Task & & en-fr & en-de & en-ru & en-zh \\
  \hline
          & P & 81.9 & 82.2 & 79.9 & 76.7 \\
   Train &  R & 69.1 & 70.1 & 67.8 & 67.1 \\
         & F1 & 74.9 & 76.1 & 73.3 & 71.6 \\
  \MC{2}{|r|}{Threshold}
              & 0.58& 0.50& 0.57& 0.64 \\
  \hline
        & P  & 84.8 & 84.1 & 81.1 & 77.7 \\ Test & R  & 68.6 & 70.7 & 67.6 & 66.4 \\
        & F1 & 75.8 & 76.9 & 73.8 & 71.6 \\
  \hline
  \hline
  \end{tabular}
  \caption{Results on the BUCC test set of our approach: Precision, Recall and F-measure (\%).
        We also provide the optimal threshold on the distance.
  }
  \label{TabBuccRes}
\end{table}

We applied our approach to all language pairs of the BUCC shared task (see
Table~\ref{TabBuccRes}). We used the embeddings from
\citep{Schwenk:2017:repl4nlp} for ru and zh, which were trained on the UN
corpus.  The only task-specific adaptation is the optimization of the
threshold on the distance in the multilingual joint space.
Our system does not match the performance of the heavily tuned VIC system, but
it is on-pair with H2 on en-fr, and outperforms all other approaches by a large
margin.
We would like to emphasize that our approach uses no additional features or
classifiers, and that we apply the same approach to all language pairs. It is
nice to see that the performance varies little for the languages.

\citet{epsana:2017:ieee_embed_mine} have also evaluated their technique on the
BUCC data, but results on the official test set are not provided.  Also, their
joint encoder uses the \textit{``news-commentary''} corpus during training.
This is likely to add an important bias since all the parallel sentences in the
BUCC corpus are from the news-commentary corpus.

Since we learn multilingual embeddings for many languages in one joint space, we
can mine for parallel data for any language pair.  As an example, we have mined
for French/German and Chinese/Russian bitexts, respectively. There are no
reference alignments to optimize the threshold for this language pair. Based on the 
experiments with the other languages, we chose a value of 0.55.
In the annex, we provide examples of extracted parallel sentences for various
values of the multilingual distance.
These examples show that our approach may wrongly align sentences which are mainly
an enumeration of named entities, numerical values, etc.
Many of these erroneous alignments could be possibly excluded by some post-processing,
e.g. comparing the number of named entities in each sentence.

\section{Experimental evaluation: improving WMT'14 En-De NMT systems}

\subsection{Baseline NMT systems}

We have performed all our experiments with the freely available
Sequence-to-Sequence PyTorch toolkit from Facebook AI
Research,\footnote{\url{https://github.com/facebookresearch/fairseq-py}} called
\texttt{fairseq-py}.
It implements a convolutional model which achieves very competitive results
\citep{Gehring:2017:fairseq_icml}.  We use this system to show the improvements
obtained by filtering the standard training data and by integrating additional
mined data.
We will freely share this data so that it can be used to train different NMT architectures.

In this work, we focus on translating from English into German using the WMT'14 data.
This task was selected for two reasons:
\begin{itemize}
  \item it is the de-facto standard to evaluate NMT systems and many comparable results are available,
	e.g. \citep{sennrich:2016:acl_bpe,chung_cho:2016:arxiv_charnmt,google:2016:arxiv_nmt,Gehring:2017:fairseq_icml,Google:2017:nips_attn_is_all};
  \item only a limited amount of parallel training data is available (4.5M sentences).
        2.1M are high quality human translations and 2.4M are crawled and aligned sentences
        (Common Crawl corpus).
\end{itemize}

As in other works, we use \textrm{newstest-2014} as test set.  However, in
order to follow the standard WMT evaluation setting, we use
\texttt{mteval-v14.pl} on untokenized hypothesis to calculate case-sensitive
BLEU scores.  Note that in some papers, BLEU is calculated with
\texttt{multi-bleu.perl} on tokenized hypothesis.
All our results are for one single system only.

We trained the \texttt{fairseq-py} system with default parameters, but a
slightly different pre- and post-processing scheme.  In particular, we
lower-case all data and use a 40k BPE vocabulary \citep{sennrich:2016:acl_bpe}.
Before scoring, the case of the hypothesis is restored using a recaser
trained on the WMT German news data.
Table~\ref{TabResBase} gives our baseline results using the provided data as it
is.  We distinguish results when training on human labeled data only, i.e.
Europarl and News Commentary (2.1M sentences), and with all WMT'14 training
data, i.e. human + Common Crawl (total of 4.5M sentences).
\citet{Gehring:2017:fairseq_icml} report a tokenized BLEU score of 25.16
on a slightly different version of newstest-2014 as defined in
\citep{luang:2015:emnlp_nmt}.\footnote{This version uses a subset of 2737 out of
3003 sentences.}
Please remember that the goal of this paper is not to set a new
state-of-the-art in NMT on this data set, but to show relative
improvement with respect to a competitive baseline.

\begin{table}[t]
  \centering
  \begin{tabular}[t]{|c||c|c|c|}
  \hline
  \MR{2}{*}{Corpus} & Human only &  All WMT'14 \\
  & (Eparl+NC) &  (Eparl+NC+CC) \\
  \hline
  \#sents & 2.1M  & 4.5M \\
  BLEU & 21.87 & 24.75 \\
  \hline
  \end{tabular}
  \caption{Our baseline results on WMT'14 en-de.}
  \label{TabResBase}
  \vspace{-5pt}
\end{table}


\subsection{Filtering Common Crawl}

The Common Crawl parallel corpus is provided by the organizers of WMT'14.  We do not know
how this corpus was produced, but like all crawled corpora, it is inherently
noisy.
To filter that corpus, we first embed all the sentences into the joint space
and calculate the cosine distance between the English source and the provided
German translation. We then extract subsets of different sizes as a function of
the threshold on this distance.

\begin{table}[h!]
  \centering
  \begin{tabular}[t]{|c||c|c|c|}
  \hline
  All & Commas & $<$50 words & LID \\
  \hline
  2399k & 2144k & 2071k & 1935k \\
  \hline
  \end{tabular}
  \caption{Pre-processing of the Common Crawl corpus before distance-based filtering.}
  \label{TabCCPre}
  \vspace{-5pt}
\end{table}

After some initial experiments,  it turned out that some additional steps are
needed before calculating the distances (see Table~\ref{TabCCPre}):
1) remove sentences with more than 3 commas. Those are indeed often enumerations of names, cities, etc.
   While such sentences maybe useful to train NMT systems, the multilingual distance is not very reliable to distinguish list of named entities;
2) limit to sentences with less than 50 words; 
3) perform language identification (LID) on source and target sentences.
These steps discarded overall 19\% of the data. It is surprising that almost 6\% of the
data seems to have the wrong source or target language.\footnote{LID itself may
also commit errors, we used\\ \url{https://fasttext.cc/docs/en/language-identification.html}}

\begin{figure}[t!]
  \vspace{-10.5cm}
  \centering
  \hspace*{-1cm}\includegraphics[width=0.8\textwidth]{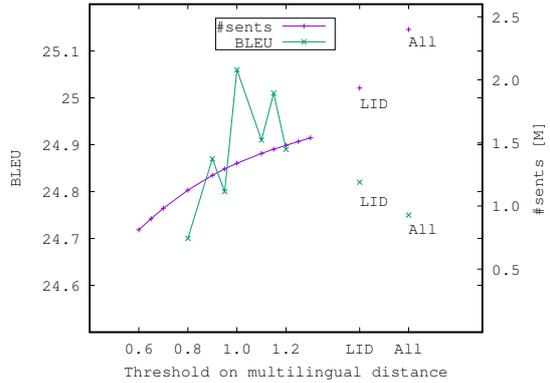}
  \vspace*{-1.8cm}
  \caption{Filtering the Common Crawl corpus:
	size of corpus (pink) and BLEU scores (green).}
  \label{FigCCfilt}
  \vspace{-10pt}
\end{figure}

Figure~\ref{FigCCfilt} (pink curve) shows the amount of data as a function of
the threshold on the multilingual distance. Some human inspection of the
filtered corpus indicated that the translations start to be wrong for a
threshold larger than 1.0.  Therefore, we build NMT systems using a filtered
version of Common Crawl for thresholds in the range of 0.8 to 1.2 (see
Figure~\ref{FigCCfilt}, green curve).  It is good to see that the BLEU score
increases when less but better data is used and then decreases again since we
discard too much data.  Best performance of 25.06 BLEU is achieved for a
threshold of 1.0.  This corresponds to a gain of 0.3 BLEU on top of a very
competitive baseline (24.75$\rightarrow$25.06), using only 3.4M instead of the
original 4.5M sentence pairs. We actually discard almost half of the Common
Crawl data.
For comparison, we also trained an NMT system using the pre-processed Common
Crawl corpus of 1.9M sentences (cf. Table~\ref{TabCCPre}), but without
distance-based filtering. This gives a BLEU score of 24.82, a small 0.07
change.

Aiming at a compromise between speed and full convergence, we trained all
systems for 55 epochs which takes less than two days on 8 NVidia GPU100s.
Longer training may improve the overall results.

\subsection{Mining Parallel Data in WMT News}

In the framework of the WMT evaluation, large news corpora are provided: 144M
English and 187M German sentences (after removing sentence with more than 50
words).  As in section~\ref{SectBucc}, we embed all sentences into the joint
space.  For each source sentence, we search for the $k$-nearest sentences in
the target language.  We use $k=20$ since it can happen that for the same
source sentence, several possible translations are found (different news sites
reporting on the same fact with different wordings).  This search has a
complexity of $O(N\times M)$, while filtering presumed parallel corpora is
$O(N)$.  In our case, $144M \times 185M$ amounts to $2.7 \times 10^{16}$ distance
calculations.  This can be quite efficiently done with the highly optimized
FAISS toolkit \citep{FAISS:2017:arxiv}.

\begin{table}[b!]
  \centering
  \begin{tabular}{|r|r||*{3}{@{\,}c@{\,}|}}
    \hline
    & & \MC{3}{c|}{BLEU} \\
    \MR{2}{*}{Threshold}
     & \MR{2}{*}{\#Sents} & Mined & Eparl & All \\
    &  & alone & + mined & + mined \\
    \hline
    \hline
    baseline & -&  -   & 21.87 & 25.06 \\
    \hline
    0.25 & 1.0M & 4.18 & \bf 22.32 & \bf 25.07 \\
    0.26 & 1.5M & 5.17 & 22.09 & - \\
    0.27 & 1.9M & 5.92 & 21.97 & - \\
    0.28 & 2.5M & 6.48 & 22.29 & 25.03 \\
    0.29 & 3.3M & 6.01 & 22.10 & - \\
    0.30 & 4.3M & 7.77 & 22.24 & - \\
    \hline
  \end{tabular}
  \caption{BLEU scores when training on the mined data only, adding it (at different thresholds)
	to the human translated training corpus (Eparl+NC) and to our best system using filtered
	Common Crawl.}
  \label{TabResMined}
\end{table}

To start, we trained NMT systems on the extracted data only (see
Table~\ref{TabResMined}, 3rd column).  As with the Common Crawl corpus, we
discarded sentences pairs with the wrong language and many commas.  By
varying the threshold on the distance between two sentences in the embedding
space, we can extract various amounts of data. However, the larger the
threshold, the more unlikely the sentences are translations.  Training on 1M
mined sentences gives a modest BLEU score of 4.18, which increases up to 7.77
when 4.3M sentences are extracted.  This result is well below an NMT system
trained on \textit{``real parallel data''}.

\begin{figure}[t!]
  \vspace{-10.5cm}
  \centering
  \hspace*{-1cm}\includegraphics[width=0.8\textwidth]{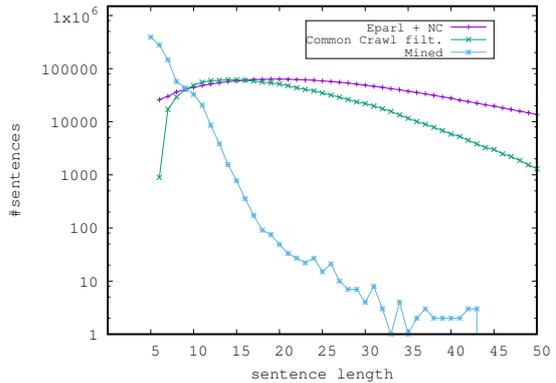}
  \vspace*{-1.8cm}
  \caption{Number of sentences as a function of their length, for
        WMT'14 training corpora and the mined news texts.}
  \label{FigMinedHisto}
  \vspace{-10pt}
\end{figure}

We have observed that the length distribution of the mined sentences is very
different of the one of the WMT'14 training corpora (see
Figure~\ref{FigMinedHisto}).  The average sentence length for all the WMT
training corpora is 24, while it is only 8 words for our mined texts.  On one
hand, it could be of course that our distance based mining approach works badly
for long sentences.  But on the other hand, the longer the sentences, the more
unlikely it is to find perfect translation in crawled news data.  If we shuffle
the Europarl corpus and consider it as a comparable corpus, our approach is
able to extract more than 95\% of the translation pairs.  It is also an open
question how short sentences impact the training of NMT systems.
Further research in those directions is needed.

When adding our mined data to the Europarl and News Commentary corpora (2.1M
sentences), we are able to achieve an improvement of 0.45 BLEU
(21.87$\rightarrow$22.32, 4th column of Table~\ref{TabResMined}).
However, we observe no improvement when adding the mined data to our best system
which uses the filtered Common Crawl data (5th column of Table~\ref{TabResMined}).
It could be that some of our mined data is actually a subset of Common Crawl.

\section{Conclusion}

We have shown that a simple cosine distance in a joint multilingual sentence
embedding space can be used to filter noisy parallel data and to mine for bitexts
in large news collections.
We were able to improve a competitive baseline on the WMT'14 English
to German task by 0.3 BLEU by filtering out 25\% of the training data.
We will make the filtered and extracted data freely available, as well
as a tool to filter noisy bitexts in nine languages.

There are many directions to extend this research, in particular to scale-up to
larger corpora. We will apply it to the data mined by the European ParaCrawl
project.\footnote{\url{http://paracrawl.eu/download.html}}
The proposed multilingual sentence distance could
be also used in MT confidence estimation, or to filter
back-translations of monolingual data \citep{sennrich:2016:acl_mono}.

\section*{Acknowledgments}

We would like to thank Matthijs Douze from Facebook AI research for many
helpful comments on the use of FAISS.

\bibliographystyle{acl_natbib}
\bibliography{strings_short,mt_hs}

\newpage
\clearpage

\begin{table*}[!t]
  \vspace*{-0.9in}
  \hspace*{-1in}
  \includegraphics[width=1.30\textwidth]{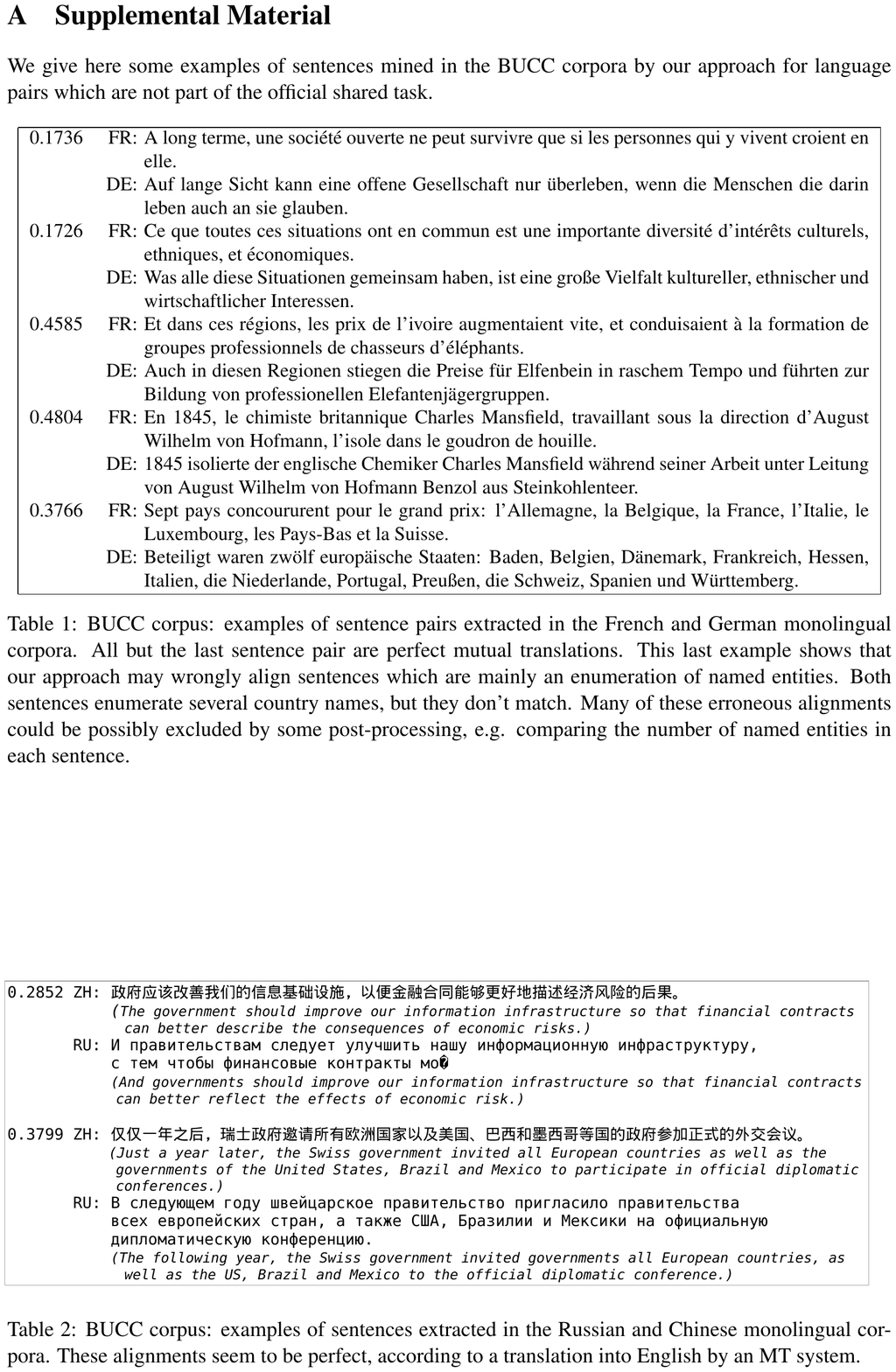}
\end{table*}

\end{document}